%% file: main.tex
\documentstyle[11pt,twocolumn]{article}


\title{\bf Applying a Hybrid Query Translation Method to
Japanese/English Cross-Language Patent Retrieval}

\author{\Large Masatoshi Fukui$^{\dagger}$~~Shigeto
Higuchi$^{\dagger}$~~Youichi Nakatani$^{\dagger}$~~\medskip \\
\Large Masao Tanaka$^{\dagger}$~~Atsushi Fujii$^{\ddagger}$~~Tetsuya
Ishikawa$^{\ddagger}$}

\date{$^\dagger$Japan Patent Information Organization \\
Satoh Daiya Bldg., 1-7 Toyo 4-Chome Koto-ku 135-0016, JAPAN \medskip \\
$^\ddagger$University of Library and Information Science \\
1-2 Kasuga Tsukuba 305-8550, JAPAN \\ \smallskip
\smallskip {\tt E-mail:~fujii@ulis.ac.jp}}

\newcommand{\eq}[1]{(\ref{#1})}

\input{psfig.tex}

\begin{document}

\maketitle\thispagestyle{empty}

\begin{abstract}
  This paper applies an existing query translation method to
  cross-language patent retrieval. In our method, multiple
  dictionaries are used to derive all possible translations for an
  input query, and collocational statistics are used to resolve
  translation ambiguity. We used Japanese/English parallel patent
  abstracts to perform comparative experiments, where our method
  outperformed a simple dictionary-based query translation method, and
  achieved 76\% of monolingual retrieval in terms of average
  precision.
\end{abstract}

\section{Introduction}
\label{sec:introduction}

Since 1978, JAPIO (Japan Patent Information Organization) has operated
PATOLIS, which is one of the first on-line patent retrieval services
in Japan, and currently provides clients (i.e., 8,000 Japanese
companies) with patent information from 62 countries and 5
international organizations. At the same time, since a patent obtained
in a single country can be protected in multiple countries
simultaneously, it is feasible that users are interested in retrieving
patent information across languages. Motivated by this background,
JAPIO manually summarizes each patent document submitted in Japan into
approximately 400 characters, and translates the summarized documents
into English, which are provided on PAJ (Patent Abstract of Japan)
CD-ROMs\footnote{Copyright by Japan Patent Office.}.

In this paper, we target cross-language information retrieval (CLIR)
in the context of patent retrieval, and evaluate its effectiveness
using Japanese/English patent abstracts on PAJ CD-ROMs.

In brief, existing CLIR systems are classified into three approaches:
(a) translating queries into the document
language~\cite{ballesteros:sigir-98,davis:sigir-97}, (b) translating
documents into the query language~\cite{mccarley:acl-99,oard:amta-98},
and (c) representing both queries and documents in a
language-independent
space~\cite{carbonell:ijcai-97,gonzalo:chum-98,littman:clir-98,salton:jasis-70}.
However, since developing a CLIR system is expensive, we used the CLIR
system proposed by Fujii and
Ishikawa~\cite{fujii:ntcir-99,fujii:emnlp-vlc-99}, which follows the
first approach.

This system has partially been developed for the NACSIS test
collection~\cite{kando:sigir-99}, which consists of 39 Japanese
queries and approximately 330,000 technical abstracts in Japanese and
English.  However, since patent information usually includes technical
terms, it is expected that this system also will perform reasonably
for patent abstracts.

\section{System Description}
\label{sec:system}

\begin{figure}[t]
  \begin{center}
    \leavevmode
    \psfig{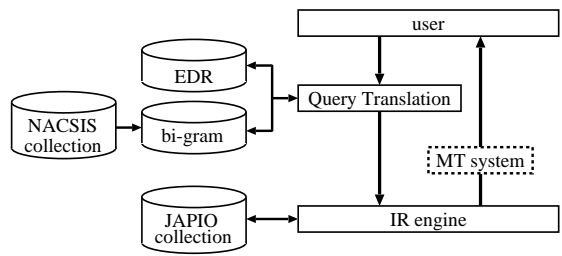}
  \end{center}
  \caption{The overall design of our cross-language patent retrieval
    system.}
  \label{fig:system}
\end{figure}

Figure~\ref{fig:system} depicts the overall design of our CLIR system,
in which we combine a query translation module and an IR engine for
monolingual retrieval.  Unlike the original system proposed by Fujii
and Ishikawa~\cite{fujii:ntcir-99,fujii:emnlp-vlc-99} targeting the
NACSIS collection, we use the JAPIO collection for the target
documents. Here, the JAPIO collection is a subset of PAJ CD-ROMs. We
will elaborate on this collection in
Section~\ref{sec:experimentation}. In this section, we briefly explain
the retrieval process based on Figure~\ref{fig:system}.

First, query translation is performed for the source language query to
output the translation. For this purpose, a hybrid method integrating
multiple resources is used. To put it more precisely, the EDR
technical/general dictionaries~\cite{edr:95} are used to derive all
possible translation candidates for words and phrases included in the
source query. In addition, for words unlisted in dictionaries,
transliteration is performed to identify phonetic equivalents in the
target language.

Then, bi-gram statistics extracted from NACSIS documents in the target
language are used to resolve the translation ambiguity. Ideally,
bi-gram statistics should be extracted from the JAPIO
collection. However, since the number of documents in this collection
is relatively small, when compared with the NACSIS collection (see
Section~\ref{sec:experimentation}), we avoided the data sparseness
problem.

Since our system is bidirectional between Japanese and English, we
tokenize documents with different methods, depending on their
language. For English documents, the tokenization involves eliminating
stopwords and identifying root forms for inflected content words. For
this purpose, we use WordNet~\cite{fellbaum:wordnet-98}, which
contains a stopword list and correspondences between inflected words
and their root form.

On the other hand, we segment Japanese documents into lexical units
using the ChaSen morphological analyzer~\cite{matsumoto:chasen-97},
which has commonly been used for much Japanese NLP research, and
extract content words based on their part-of-speech information.

Second, the IR engine searches the JAPIO collection for documents
relevant to the translated query, and sorts them according to the
degree of relevance, in descending order. Our IR engine is based on
the vector space model, in which the similarity between the query and
each document (i.e., the degree of relevance of each document) is
computed as the cosine of the angle between their associated
vectors. We use the notion of TF$\cdot$IDF for term weighting. Among a
number of variations of term weighting
methods~\cite{salton:ipm-88,zobel:sigir-forum-98}, we tentatively use
the formulae as shown in Equation~\eq{eq:tf_idf}.
\begin{equation}
  \label{eq:tf_idf}
  \begin{array}{lll}
    TF & = & 1 + \log(f_{t,d}) \\
    \noalign{\vskip 1.2ex}
    IDF & = & \log(\frac{\textstyle N}{\textstyle n_{t}})
  \end{array}
\end{equation}
Here, $f_{t,d}$ denotes the frequency that term $t$ appears in
document $d$, and $n_{t}$ denotes the number of documents containing
term $t$. $N$ is the total number of documents in the collection.

For the indexing process, we first tokenize documents as explained
above (i.e., we use WordNet and ChaSen for English and Japanese
documents, respectively), and then conduct the word-based
indexing. That is, we use each content word as a single indexing term.

Finally, since retrieved documents are not in the user's native
language, we optionally use a machine translation system to enhance
readability of retrieved documents.

\section{Experimentation}
\label{sec:experimentation}

Since no test collection for Japanese/English patent retrieval is
available to the public, we produced our test collection (i.e., the
JAPIO collection), which consists of three Japanese queries and
Japanese/English comparable abstracts.

Each query, which was manually produced, consists of the description
and narrative, and corresponds to different domains, i.e., electrical
engineering, mechanical engineering and chemistry.
Figure~\ref{fig:query} shows the three query descriptions in the
second column.

\begin{figure*}[htbp]
  \begin{center}
    \leavevmode
    \small
    \begin{tabular}{llrr} \hline\hline
      {\hfill\centering IPC\hfill} & {\hfill\centering
      Description\hfill} & \#Relevant & \#Documents \\ \hline
      electronics & GPS car navigation system based on VICS & 930 &
      7,526 \\
      mechanics & eliminating dioxin in burning solid wastes & 451 &
      8,214 \\
      chemistry & antibacterial plastic combining inorganic
      materials & 473 & 5,902 \\
      \hline
    \end{tabular}
    \caption{Query descriptions in the JAPIO collection.}
    \label{fig:query}
  \end{center}
\end{figure*}

In conventional test collections, relevance assessment is usually
performed based on the pooling method~\cite{voorhees:sigir-98}, which
first pools candidates for relevant documents using multiple retrieval
systems. However, since in our case only one system described in
Section~\ref{sec:system} is currently available, a different
production method was needed.

To put it more precisely, for each query (domain), target documents
were first collected based on the IPC classification number, from PAJ
CD-ROMs in 1993-1998. Then, for each query, three professional human
searchers, who were allowed to enhance queries based on thesauri and
their introspection, searched the target documents for relevant
documents.

Thus, in practice, the JAPIO collection consists of three different
document collections corresponding to each query.  In
Figure~\ref{fig:query}, the third and fourth columns denote the number
of relevant documents and the total number of target documents for
each query.

We compared the following methods:
\begin{itemize}
\item Japanese-English CLIR, where all possible translations derived
  from EDR dictionaries and the transliteration method were used as
  query terms (JEALL),
\item Japanese-English CLIR, where disambiguation based on bi-gram
  statistics were performed, and $k$-best translations were used as
  query terms (JEDIS),
\item Japanese-Japanese monolingual IR (JJ).
\end{itemize}
Here, we empirically set \mbox{$k=1$}. Although the performance of
JEDIS did not significantly differ as long as we set a small value of
$k$ (e.g., \mbox{$k=5$}), we achieved the best performance when we set
\mbox{$k=1$}.

\begin{figure}[t]
  \begin{center}
    \leavevmode
    \psfig{file=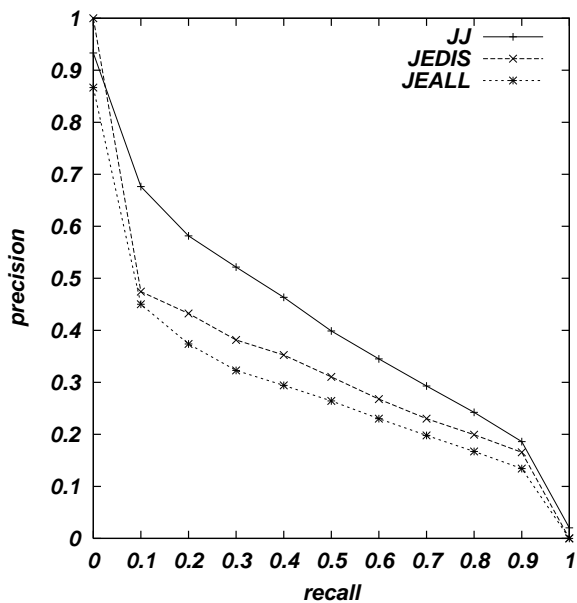,height=3.2in}
  \end{center}
  \caption{Recall-precision curves for different methods.}
  \label{fig:rp}
\end{figure}

Figure~\ref{fig:rp} shows recall-precision curves for the above three
methods, where JEDIS generally outperformed JEALL, and JJ generally
outperformed both JEALL and JEDIS, regardless of the recall.  The
difference between JEALL and JEDIS is attributed to the fact that
JEDIS resolved translation ambiguity based on bi-gram statistics
extracted from the NACSIS collection. Thus, we can conclude that the
use of bi-gram statistics (even extracted from a collection other than
the JAPIO collection) was effective for the query translation.

Table~\ref{tab:avg_pre} shows the non-interpolated average precision
values, averaged over the three queries, for each method.  This table
shows that JJ outperformed JEALL and JEDIS, JEDIS outperformed JEALL,
and the average precision value for JEDIS was 76\% of that obtained
with JJ.

These results are also observable in existing CLIR experiments using
the TREC and NACSIS collections. Thus, we conclude that our
cross-language patent retrieval system is relatively comparable with
those for newspaper articles and technical abstracts in performance.

However, we could not conduct statistical testing, which investigates
whether the difference in average precision is meaningful or simply
due to chance~\cite{hull:sigir-93}, because the number of queries is
small. We concede that experiments using a larger number of queries
need to be further explored.

\section{Conclusion}
\label{sec:conclusion}

In this paper, we explored Japanese/English cross-language patent
retrieval. For this purpose, we used an existing cross-language IR
system relying on a hybrid query translation method, and evaluated its
effectiveness using Japanese queries and English patent abstracts.
The experimental results paralleled existing experiments. That is, we
found that resolving translation ambiguity was effective for the query
translation, and that the average precision value for cross-language
IR was approximately 76\% of that obtained with monolingual IR.
Future work will include qualitative/quantitative analyses based on a
larger number of queries.

\begin{table}[htbp]
  \begin{center}
    \caption{Non-interpolated average precision values,
    averaged over the three queries, for different methods.}
    \medskip
    \leavevmode
    \small
    \begin{tabular}{lcc} \hline\hline
      Method & Avg. Precision & Ratio to JJ \\ \hline
      JJ & 0.4151 & -- \\
      JEDIS & 0.3156 & 0.7603 \\
      JEALL & 0.2709 & 0.6526 \\
      \hline
    \end{tabular}
    \label{tab:avg_pre}
  \end{center}
\end{table}

\small

\bibliographystyle{jplain}

\end{document}